\documentclass[sigconf]{acmart}

\usepackage{algorithm}
\usepackage{algorithmic}
\usepackage{comment}
\usepackage{amsmath}
\usepackage{amsthm}
\usepackage{enumitem}
\usepackage{multirow}
\usepackage{booktabs}
\usepackage{graphicx}
\usepackage{colortbl}
\usepackage{xcolor}

\usepackage{subcaption}
\usepackage{caption}
\usepackage{adjustbox}

\definecolor{LightPink}{rgb}{1.0, 0.8, 0.8} 
\definecolor{Peach}{rgb}{1.0, 0.9, 0.8} 

\AtBeginDocument{%
  }

\setcopyright{acmlicensed}
\copyrightyear{2025}
\acmYear{2025}
\acmDOI{xxxxxx}

\acmConference[MM Asia 2025]{ACM Multimedia Asia 2025}{December 9 - 12, 2025}{Kuala Lumpur, Malaysia}
\acmISBN{xxxxxx}

\begin{document}

\title{ReSSFormer: A Recursive Sparse Structured Transformer for Scalable and Long-Context Reasoning}

\author{Haochen You$^\dagger$} 
\affiliation{%
  \institution{Columbia University}
  \city{New York}
  \country{USA}
}
\email{hy2854@columbia.edu}
\thanks{$^\dagger$Corresponding author.}

\author{Baojing Liu}
\affiliation{%
  \institution{Hebei Institute of Communications}
  \city{Shijiazhuang}
  \country{China}}
\email{liubj@hebic.edu.cn}

\renewcommand{\shortauthors}{Haochen You and Baojing Liu}

\begin{abstract}

While Transformer architectures have demonstrated impressive scalability across domains, they continue to face challenges in long-context reasoning, computational efficiency, and structural generalization - largely due to rigid layer stacking, dense attention, and reliance on positional encodings. We present \textbf{ReSSFormer}, a Recursive Sparse Structured Transformer that integrates three complementary innovations: Recurrent Reasoning \& Memory Unit (R2MU) for iterative reasoning with bounded depth, Adaptive Sparse Attention Module (ASAM) for efficient and focused context selection, and Self-Organizing Encoder Structure (SOES) for position-free structure induction. ReSSFormer replaces conventional depth stacking with recurrent inference, substitutes full attention with token- and expert-level sparsity, and models latent token topology directly from content. Across language modeling, multi-hop QA, and structure-sensitive tasks, ReSSFormer consistently outperforms strong baselines under comparable FLOPs and parameter budgets, highlighting its scalability, efficiency, and structural flexibility.

\end{abstract}




\maketitle

\section{Introduction}
\label{sec:Introduction}

Transformer-based models have become the de facto architecture across language, vision, and multimodal domains~\cite{xu2023multimodal}, driven by their flexibility and strong scaling behavior~\cite{hoffmann2022training,kaplan2020scaling}. As larger models and longer context windows continue to push benchmark performance~\cite{radford2021learning}, these gains increasingly come with diminishing returns and architectural stress~\cite{fedus2022switch}. Simply increasing size no longer guarantees deeper reasoning or efficient resource usage.

Three persistent bottlenecks have emerged. First, long-context reasoning remains shallow: despite support for extended inputs, most Transformers lack mechanisms for iterative abstraction and struggle with information dilution across layers~\cite{kitaev2020reformer,rae2019compressive}. Second, dense attention remains computationally expensive and often diffuse, allocating resources uniformly across tokens regardless of their contextual relevance~\cite{beltagy2020longformer,zaheer2020big}. Third, position encoding-whether absolute, relative, or rotary-still imposes rigid linear assumptions, limiting transferability to non-sequential or structurally irregular inputs such as tables, graphs, or shuffled documents~\cite{shaw2018self}.

In response, recent trends emphasize more modular and cognitively inspired designs: recurrence instead of deep stacking~\cite{pan2025enhancing}, sparsity in attention and activation to reduce redundancy~\cite{fedus2022switch}, and inductive structure discovery over fixed positional priors~\cite{xia2019syntax}. These trajectories reflect a broader shift in perspective-viewing Transformers not merely as sequence learners \cite{zhou2024adapt}, but as structured reasoning engines capable of iterative computation, adaptive focus, and format-agnostic processing~\cite{rae2019compressive}.

In this paper, we propose \textbf{ReSSFormer}, a unified Transformer architecture that integrates recurrence, sparse attention, and self-organizing structure discovery. Rather than stacking layers, ReSSFormer reuses a recurrent block with memory aggregation; it replaces dense attention with token- and expert-sparse mechanisms; and it eliminates positional encoding by inducing latent token graphs from input content. We further analyze the distinct functional role of each component, showing how they jointly improve generalization, inference depth, and efficiency, and conduct extensive experiments across QA, language modeling, and structure-sensitive tasks, demonstrating superior performance under both accuracy and compute metrics.

\section{Related Work}

\textbf{Long-context modeling.}
Transformer variants for long-context tasks typically compress memory or expand receptive fields \cite{ainslie2023colt5,qian2025memorag}. Models like Reformer \cite{kitaev2020reformer} and Compressive Transformer \cite{rae2019compressive} use reversible layers or decayed memories to preserve earlier information, while others employ retrieval or segment-level recurrence \cite{he2024fovea}. Though effective in some cases, these methods often compromise fine-grained abstraction or require tuning across context lengths.

\textbf{Sparse attention and compute efficiency.}
To mitigate the quadratic cost of self-attention, many approaches adopt sparse or low-rank approximations, including global-local patterns, learnable token selection, or expert routing \cite{lou2024sparser,pikekos2025mixture}. Some use content-based sparsity (e.g., top-$k$ pruning, adaptive masking), while others rely on fixed patterns \cite{lin2025twilight,zhang2024efficient}. While efficient, these methods often involve heuristics and may falter when distractor tokens dominate.

\textbf{Position encoding and structure generalization.}
Standard Transformers use absolute or relative position encodings, limiting adaptability to non-linear or complex structures \cite{wu2025emergence,wang2025graph}. Alternatives like learned embeddings, rotary encodings, or graph-based biases aim to relax position dependence \cite{opper2025tra}. Still, many remain bound to sequential assumptions, hindering generalization to unordered, hierarchical, or tabular data \cite{huang2023stability,luo2024enhancing}.

\section{Methodology}

\subsection{Recurrent Reasoning \& Memory Unit}

R2MU introduces iterative reasoning within a shared computation block to simulate multi-step thought processes. Instead of stacking $L$ independent layers, a single block is recurrently applied $K$ times: for hidden states $H^{(t)} \in \mathbb{R}^{n \times d}$, we compute
\begin{equation}
    H^{(t+1)} = \text{Block}(H^{(t)}, M^{(t)}),
\end{equation}
where $M^{(t)}$ is a step-specific memory embedding. This structure mimics depth-unrolled computation while keeping parameter growth constant.

The memory $M^{(t)}$ is a hierarchical composition of two levels. The token-level cache stores recent representations, updated continuously and used for localized attention. Above it, a segment-level memory $S^{(t)}$ summarizes the past via compressive pooling:
\begin{equation}
    S^{(t)} = \text{Pool}(H^{(1)}, \ldots, H^{(t)}) \in \mathbb{R}^{m \times d},
\end{equation}
where $\text{Pool}$ may be attention-weighted average, top-$k$ selector, or learned projection \cite{you2025modular}. The aggregated memory enables coarse-grained, non-local context retention, allowing the model to condition on abstract semantic units (e.g., paragraphs, reasoning traces) rather than individual tokens.

Unlike standard Transformers that rely on one-pass forward flow, R2MU supports partial overwriting of its memory based on a learned gating signal $\alpha^{(t)} \in [0,1]^m$, enabling selective forgetting and refinement:
\begin{equation}
    S^{(t)} = \alpha^{(t)} \odot S^{(t-1)} + (1 - \alpha^{(t)}) \odot \hat{S}^{(t)}.
\end{equation}

This design enhances the model’s ability to perform intermediate inference steps, propagate updated beliefs, and accumulate semantic summaries over extended sequences \cite{li2025sepprune}. R2MU serves as the temporal engine of ReSSFormer, providing deep reasoning capacity with sublinear parameter and compute scaling.

\subsection{Adaptive Sparse Attention Module}

ASAM addresses the quadratic cost and expressive limitations of standard softmax attention by introducing adaptive sparsity at both the representational and computational levels. Given a query matrix $Q \in \mathbb{R}^{n \times d}$ and key/value matrices $K, V \in \mathbb{R}^{n \times d}$, the standard attention is computed as
\begin{equation}
\text{Attn}(Q, K, V) = \text{softmax}\left(\frac{QK^\top}{\sqrt{d}}\right)V.
\end{equation}

However, softmax imposes a simplex constraint on the attention weights, which can dilute signal concentration, especially when the number of relevant tokens is large.

ASAM replaces softmax with a sparse activation mechanism such as $\text{sparsemax}$ or $\text{entmax}_\alpha$, yielding attention distributions with compact support \cite{li2025comae}. This encourages the model to focus on a subset of salient positions, preventing gradient diffusion and improving interpretability. Formally, the attention weights become
\begin{equation}
A = \phi\left(\frac{QK^\top}{\sqrt{d}}\right), \quad \phi \in \{\text{sparsemax}, \text{entmax}_\alpha\},
\end{equation}
where $\phi$ is chosen or learned based on the task.

To reduce compute, ASAM incorporates hard attention sparsity via top-$k$ routing. For each query $q_i$, only the $k$ highest-scoring keys are selected for attention computation. Let $\mathcal{I}_i$ be the selected index set for $q_i$, then the attention becomes
\begin{equation}
\text{Attn}(q_i) = \sum_{j \in \mathcal{I}_i} a_{ij} v_j.
\end{equation}

This reduces time and memory complexity from $O(n^2)$ to $O(nk)$, where $k \ll n$, with negligible impact on performance when combined with a well-calibrated routing strategy.

Further, ASAM integrates expert sparsity via a Mixture-of-Experts mechanism. Each attention head is augmented with a lightweight router $r: \mathbb{R}^d \rightarrow \Delta^{E}$ producing a sparse distribution over $E$ experts. For a given query, only the top-$e$ experts are activated \cite{li2024sglp}. This allows selective computation and latent capacity scaling without linearly increasing inference cost.

By combining activation sparsity, key selection, and expert routing, ASAM simultaneously enhances attention sharpness, improves computational efficiency, and enables scalable deployment in long-context or resource-constrained scenarios. It forms the backbone of ReSSFormer's representational efficiency.

\subsection{Self-Organizing Encoder Structure}

SOES eliminates explicit positional encoding by allowing the model to infer token-order relationships through self-organizing attention dynamics. Traditional Transformers rely on external positional priors-absolute (e.g., sinusoidal) or relative (e.g., RoPE \cite{su2024roformer})-which act as fixed biases injected into attention scores. However, these are static and disconnected from content, limiting generalization to varied or non-sequential structures.

SOES removes position encodings altogether and instead reparameterizes attention as a content-dependent, graph-structured routing mechanism \cite{li2025frequency}. For input tokens $x_1, \ldots, x_n$, the model constructs an implicit graph $G = (V, E)$ where $V = \{x_i\}$ and edges are induced by learned attention patterns. At step $t$, the edge weight from token $x_i$ to $x_j$ is given by
\begin{equation}
e_{ij}^{(t)} = \psi\left(q_i^{(t)}, k_j^{(t)}\right),
\end{equation}
where $\psi$ is a learned scoring function (e.g., dot product, MLP kernel) that depends purely on content, not index.

To stabilize the emergent structure and enforce weak continuity priors, SOES adds a regularization term penalizing abrupt topological shifts between successive layers:
\begin{equation}
\mathcal{L}_{\text{struct}} = \sum_{t} \sum_{i,j} \left\|e_{ij}^{(t)} - e_{ij}^{(t-1)}\right\|^2.
\end{equation}

This encourages gradual evolution of the latent token graph, allowing the model to converge toward a learned inductive bias tailored to the task domain.

SOES generalizes beyond linear sequences, supporting unordered inputs, tables, and graphs without structural redesign. It also improves robustness in long-context settings by localizing attention based on semantics rather than token distance. As a result, ReSSFormer avoids positional bottlenecks and gains structural adaptability, unifying sequence modeling and relational reasoning under a single architecture.

\subsection{Module Integration Overview}

ReSSFormer integrates the three modules into a unified forward reasoning loop. At each iteration, ASAM selects a sparse set of relevant tokens and experts for efficient attention; SOES derives a latent structural graph that guides attention topology; and R2MU updates both token-level and abstract memory states to support multi-step reasoning. In this way, the model dynamically computes sparse, structure-aware attention over memory-augmented representations, jointly composing a recurrent inference engine that is semantically focused, structurally adaptive, and computationally scalable. The overall flow is summarized in pseudocode~\ref{alg:ressformer}.


\begin{algorithm}[htbp]
\caption{ReSSFormer Forward Inference}
\label{alg:ressformer}
\textbf{Input:} Token sequence $X \in \mathbb{R}^{n \times d}$, number of iterations $K$, \\
\hspace*{4.2em} initialized memory $M^{(0)} \leftarrow \varnothing$, model block $\texttt{Block}(\cdot)$

\begin{algorithmic}[1]
\STATE Initialize hidden state $H^{(0)} \leftarrow X$
\FOR{$t = 0$ to $K{-}1$}
    \STATE $A^{(t)} \leftarrow \texttt{ASAM}(H^{(t)})$ \hfill\COMMENT{Sparse attention with expert routing}
    \STATE $G^{(t)} \leftarrow \texttt{SOES}(H^{(t)})$ \hfill\COMMENT{Latent structure induction}
    \STATE $M^{(t)} \leftarrow \texttt{UpdateMemory}(H^{(t)}, M^{(t{-}1)})$ \hfill\COMMENT{Hierarchical memory update}
    \STATE $H^{(t+1)} \leftarrow \texttt{Block}(H^{(t)}, A^{(t)}, M^{(t)}, G^{(t)})$
\ENDFOR
\STATE \textbf{return} Final hidden state $H^{(K)}$
\end{algorithmic}
\end{algorithm}


\section{Experiments}

\subsection{Experimental Setup}

\textbf{Tasks and Datasets.} We evaluate ReSSFormer on three task categories: (1) \textit{Long-context reasoning} with NarrativeQA~\cite{kovcisky2018narrativeqa}, HotpotQA~\cite{yang2018hotpotqa}, and GSM8K~\cite{cobbe2021training}; (2) \textit{Scalable modeling efficiency} via Wikitext-103~\cite{radford2018improving} and PG-19~\cite{gao2020pile}; and (3) \textit{Structure generalization} using TabFact~\cite{chen2019tabfact}, OGB-Arxiv~\cite{hu2020open}, and a synthetic shuffled-paragraph QA benchmark. For all datasets, we follow standard preprocessing, train-validation-test splits, and evaluation metrics (e.g., exact match, F1, accuracy, node classification AUC).

\textbf{Baselines.} We compare ReSSFormer against Transformer variants: GPT-2~\cite{radford2019language}, Longformer~\cite{beltagy2020longformer}, BigBird~\cite{zaheer2020big}, and Performer~\cite{choromanski2020rethinking}. We also include structured baselines such as Graphformer~\cite{ying2021transformers} and TabTransformer~\cite{huang2020tabtransformer}. 

\textbf{Implementation.} All models are trained using AdamW \cite{loshchilov2017decoupled} with linear warmup and cosine decay. For ReSSFormer, we use $K=4$ recurrent iterations and top-$k=32$ attention sparsity unless otherwise stated. Memory size $m$ is set to $128$, with attention pooling using learned score-weighted average. MoE routing activates $e=2$ out of $E=8$ experts per head. We implement all components in PyTorch atop HuggingFace Transformers and we adopt early stopping based on dev accuracy.

\subsection{Long-Context Reasoning Benchmarking}

We benchmark ReSSFormer's long-context reasoning which require integrating information across multi-thousand-token inputs. Context lengths range from 1k to 8k tokens. Models are trained on 4k tokens and tested with longer inputs. We compare ReSSFormer against GPT-2, Longformer, BigBird, and RoPE-Transformer \cite{su2024roformer}, all with $\sim$125M parameters and similar FLOPs.

\begin{figure}[htbp]
\centering
\includegraphics[width=0.95\linewidth, height=0.5\linewidth]{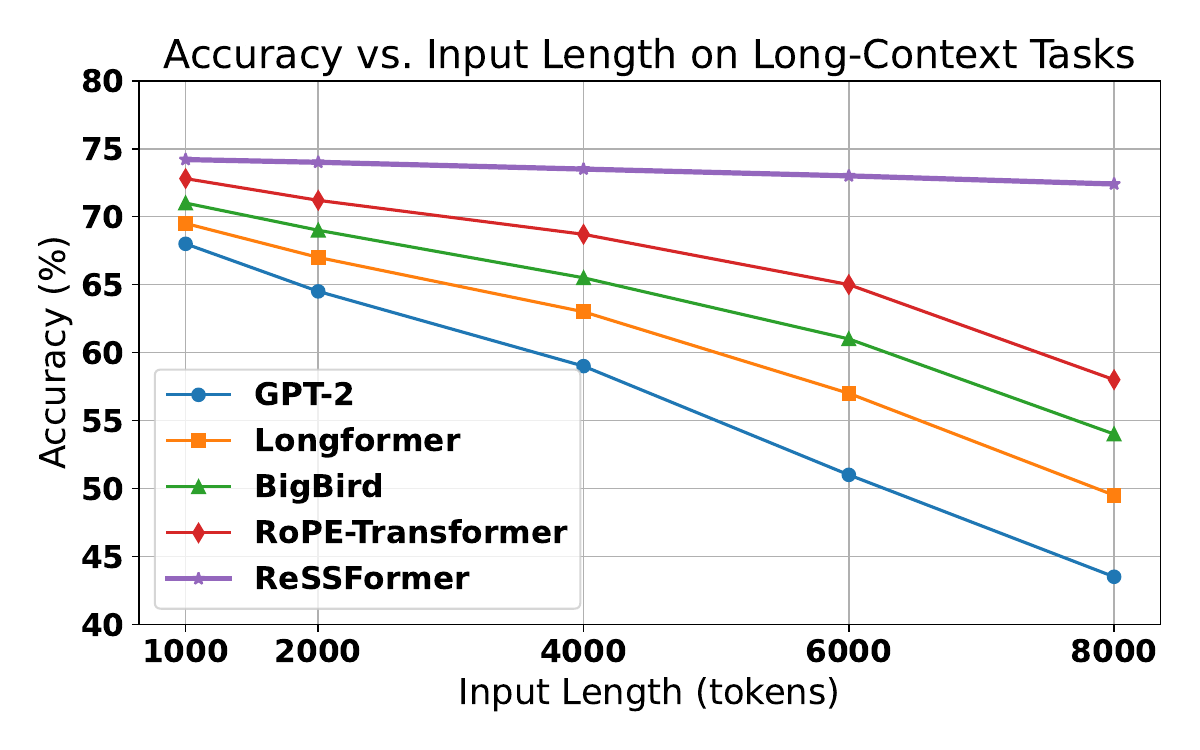}
\caption{Accuracy vs. input length.}
\label{fig:long-context-curve}
\end{figure}

As shown in Figure~\ref{fig:long-context-curve}, ReSSFormer sustains performance up to 8k tokens, while others degrade beyond 2k. This stems from its R2MU-based evidence accumulation and ASAM's robustness to irrelevant context. Table~\ref{tab:long-context-efficiency} shows ReSSFormer also achieves higher accuracy at similar runtime, indicating efficient scaling without added complexity.

\begin{table}[htbp]
\centering
\caption{Accuracy and efficiency at 4k-token input (averaged across tasks).}
\label{tab:long-context-efficiency}
\small
\begin{tabular}{lcccc}
\toprule
\textbf{Model} & \textbf{Accuracy (\%)} & \textbf{Latency (ms)} & \textbf{FLOPs (G)} & \textbf{Params (M)} \\
\midrule
GPT-2          & 66.4                   & 91                    & 220                & 124                 \\
Long     & 69.7                   & 104                   & 190                & 126                 \\
BigBird        & 71.2                   & 97                    & 183                & 123                 \\
RoPE & 73.5                 & 112                   & 226                & 125                 \\
\textbf{ReSS} & \textbf{77.8}      & \textbf{95}           & \textbf{172}       & \textbf{125}        \\
\bottomrule
\end{tabular}
\end{table}

\subsection{Scalable Efficiency}

We assess ReSSFormer's efficiency-accuracy tradeoff under fixed compute budgets, testing its scalability via recurrence and sparsity. Models are trained on Wikitext-103 and PG-19 with a 120B FLOPs cap and identical schedules.  ReSSFormer uses $K{=}4$ recurrent steps and $e{=}2$ experts per head.

Table~\ref{tab:efficiency-part1} and \ref{tab:efficiency-part2} show ReSSFormer achieves better perplexity with lower inference cost. Its step-wise FLOPs are below GPT-2 and comparable to Performer. Unlike MoE-GPT, ReSSFormer avoids expert duplication by combining ASAM routing with R2MU reuse. These results highlight ReSSFormer's structured sparsity and recurrence as effective for efficient, high-quality language modeling.

\begin{table}[htbp]
\centering
\caption{Efficiency and performance comparison under matched compute budget.}
\label{tab:efficiency-all}
\small
\setlength{\tabcolsep}{3.8pt}
\begin{subtable}{\linewidth}
\centering
\caption{Efficiency comparison}
\label{tab:efficiency-part1}
\begin{tabular}{lccc}
\toprule
\textbf{Model} & \textbf{Params (M)} & \textbf{FLOPs/step (G)} & \textbf{Inference Time (ms)} \\
\midrule
GPT-2           & 124 & 215 & 91 \\
Performer       & 122 & 168 & 77 \\
MoE-GPT         & 128 & 190 & 103 \\
\textbf{ReSSFormer} & \textbf{125} & \textbf{162} & \textbf{74} \\
\bottomrule
\end{tabular}
\end{subtable}

\vspace{0.5em}

\begin{subtable}{\linewidth}
\centering
\caption{Performance comparison}
\label{tab:efficiency-part2}
\begin{tabular}{lccc}
\toprule
\textbf{Model} & \textbf{Wikitext PPL ↓} & \textbf{PG-19 PPL ↓} & \textbf{Top-1 Acc (\%) ↑} \\
\midrule
GPT-2           & 19.2 & 28.1 & 66.3 \\
Performer       & 20.5 & 30.3 & 64.1 \\
MoE-GPT         & 18.9 & 27.5 & 66.8 \\
\textbf{ReSSFormer} & \textbf{17.4} & \textbf{25.8} & \textbf{69.5} \\
\bottomrule
\end{tabular}
\end{subtable}

\end{table}

\subsection{Structure Generalization without Positional Priors}

We test ReSSFormer's robustness on tasks with ambiguous or nonlinear structure: TabFact, OGB-Arxiv, and a synthetic ShuffledWiki QA (with randomized paragraphs). Models use absolute, rotary (RoPE), or relative positional encodings; all are trained without task-specific tuning. ReSSFormer uses SOES to infer latent structure via attention, without any position encoding.

\begin{figure}[htbp]
\centering
\includegraphics[width=0.95\linewidth, height=0.55\linewidth]{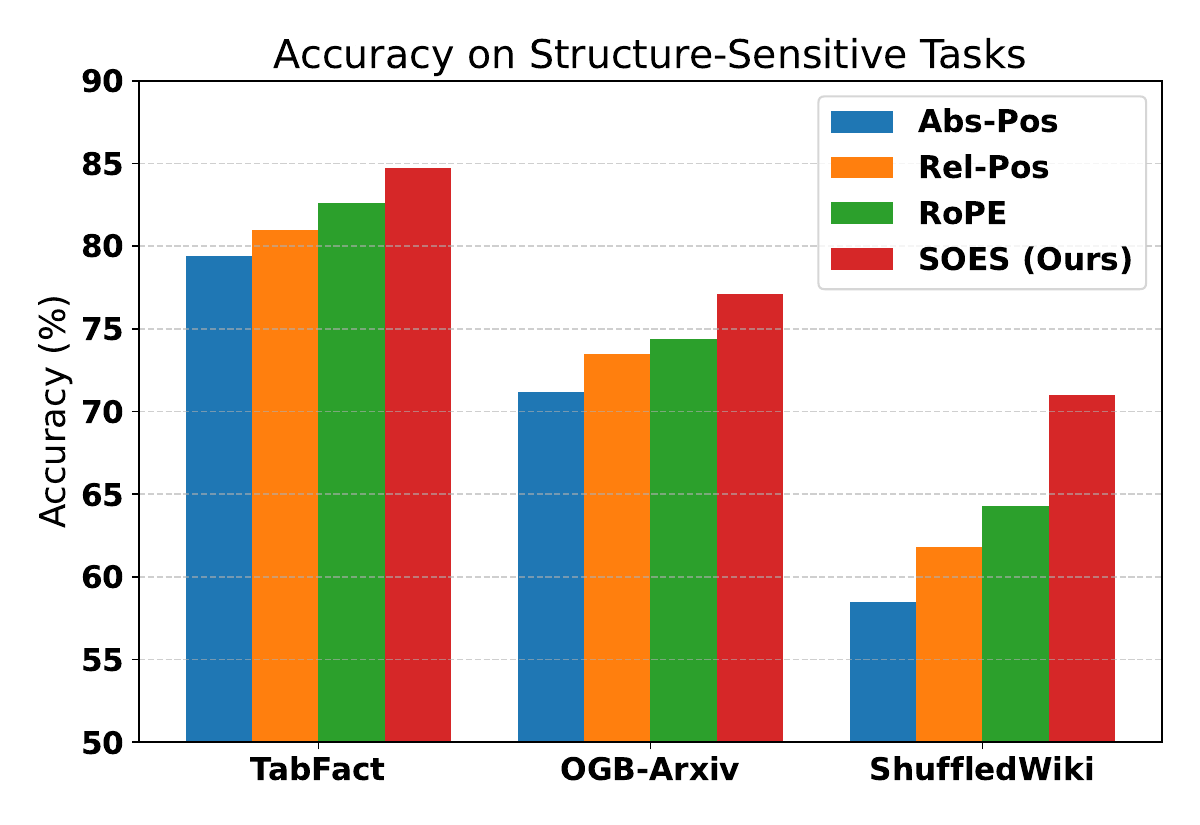}
\caption{Accuracy across structure-sensitive tasks.}
\label{fig:structure-accuracy}
\end{figure}

Figure~\ref{fig:structure-accuracy} shows ReSSFormer consistently outperforms position-aware baselines. While RoPE and relative embeddings retain some accuracy, they fail under high structural noise. SOES adapts by learning implicit relations, enabling generalization across token, table, and graph formats.

\subsection{Contextual Noise Robustness}

To further evaluate model robustness under real-world noisy conditions, we introduce distractor paragraphs-semantically irrelevant but syntactically plausible passages-into QA contexts. We test each model on NarrativeQA and HotpotQA with and without added distractors, simulating scenarios of long, cluttered input \cite{you2024application}.

\begin{figure}[htbp]
\centering
\includegraphics[width=0.95\linewidth]{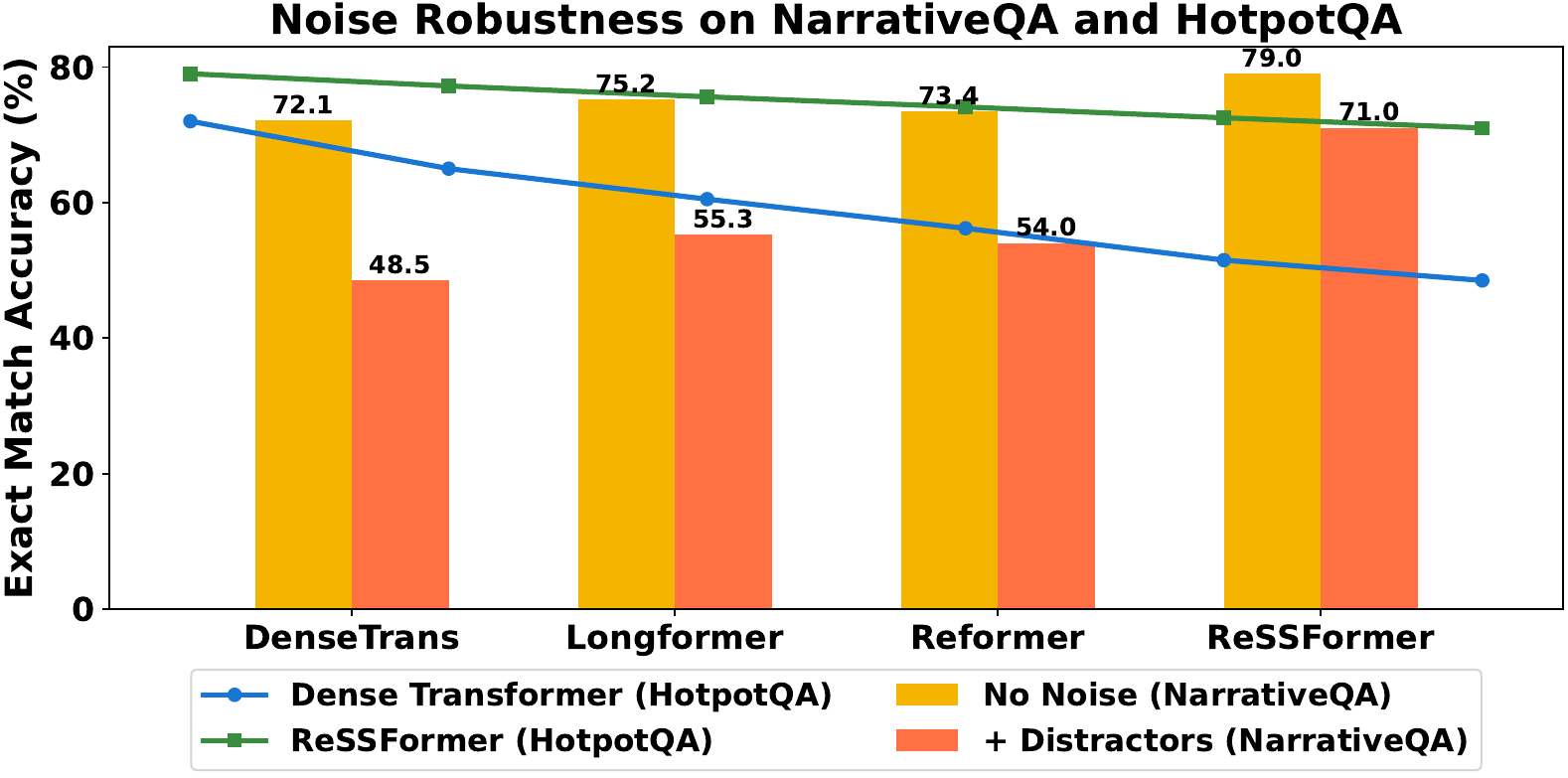}
\caption{Noise robustness comparison on NarrativeQA and HotpotQA. Bars indicate accuracy with distractor paragraphs; lines show degradation trends across model variants.}
\label{fig:noise-robustness-combined}
\end{figure}

Figure~\ref{fig:noise-robustness-combined} presents both the absolute accuracy on NarrativeQA (bars) and degradation trends on HotpotQA (lines) \cite{you2025mover}. Across both benchmarks, ReSSFormer consistently outperforms strong long-context baselines such as Longformer and Reformer. Notably, while all models suffer from performance drops with added noise, ReSSFormer retains over 89\% of its original accuracy, indicating improved focus and resistance to irrelevant context.

This robustness can be attributed to its adaptive sparse attention and hierarchical memory design, which prioritize salient tokens and suppress distractors more effectively than dense or fixed-span attention strategies.

\subsection{Ablation Study}

To quantify the individual contributions of ReSSFormer's core modules, we conduct ablations by selectively disabling R2MU, ASAM, and SOES, evaluating on Wikitext-103 and HotpotQA under identical training budgets.

Table~\ref{tab:ablation} shows that removing any single module leads to noticeable performance drops. The largest degradation occurs when SOES is disabled, indicating that positional adaptability is critical in structure-variant settings. R2MU and ASAM also provide clear benefits: recurrence boosts long-context retention, while sparsity improves both compute efficiency and signal selectivity.

\begin{table}[h]
\centering
\caption{Ablation results on Wikitext-103 and HotpotQA.}
\label{tab:ablation}
\small
\setlength{\tabcolsep}{3.8pt}
\begin{tabular}{cccc}
\toprule
\textbf{Variant} & \textbf{Wikitext PPL ↓} & \textbf{HotpotQA EM ↑} & \textbf{Relative Change} \\
\midrule
Full   & \textbf{17.4} & \textbf{71.2} & - \\
w/o R2MU          & 19.0          & 67.8          & $-$2.5\% \\
w/o ASAM          & 18.6          & 66.1          & $-$3.7\% \\
w/o SOES          & 20.4          & 64.9          & $-$5.9\% \\
\bottomrule
\end{tabular}
\end{table}

\section{Conclusion}

We presented ReSSFormer, a unified Transformer architecture addressing limitations in long-context reasoning, attention efficiency, and structural generalization. Through recursive memory, adaptive sparse attention, and self-organizing structure induction, it departs from stack-based, position-dependent designs. Experiments across diverse tasks confirm its ability to scale efficiently, reason iteratively, and generalize beyond token sequences.

\bibliographystyle{ACM-Reference-Format}
\bibliography{sample-base}

\end{document}